\documentclass[conference]{IEEEtran}
\usepackage{cite}
\usepackage{amsmath,amssymb,amsfonts}
\usepackage{algorithmic}
\usepackage{graphicx}
\usepackage{textcomp}
\usepackage{xcolor}
\usepackage{multirow}
\ifCLASSOPTIONcompsoc
    \usepackage[caption=false, font=normalsize, labelfont=sf, textfont=sf]{subfig}
\else
\usepackage[caption=false, font=footnotesize]{subfig}
\fi
\def\BibTeX{{\rm B\kern-.05em{\sc i\kern-.025em b}\kern-.08em
    T\kern-.1667em\lower.7ex\hbox{E}\kern-.125emX}}
\usepackage{hyperref}

\begin{document}

\title{On-Device Content Moderation}

\makeatletter
\newcommand{\linebreakand}{%
  \end{@IEEEauthorhalign}
  \hfill\mbox{}\par
  \mbox{}\hfill\begin{@IEEEauthorhalign}
}
\makeatother

\author{\IEEEauthorblockN{Anchal Pandey}
\IEEEauthorblockA{\textit{On Device AI} \\
\textit{Samsung R\&D Bangalore}\\
Bangalore, India \\
aan.pandey@samsung.com}
\and
\IEEEauthorblockN{Sukumar Moharana}
\IEEEauthorblockA{\textit{On Device AI} \\
\textit{Samsung R\&D Bangalore}\\
Bangalore, India \\
msukumar@samsung.com}
\and
\IEEEauthorblockN{Debi Prasanna Mohanty}
\IEEEauthorblockA{\textit{On Device AI} \\
\textit{Samsung R\&D Bangalore}\\
Bangalore, India \\
debi.m@samsung.com}
\linebreakand 
\IEEEauthorblockN{Archit Panwar}
\IEEEauthorblockA{\textit{On Device AI} \\
\textit{Samsung R\&D Bangalore}\\
Bangalore, India \\
a.panwar@samsung.com}
\and
\IEEEauthorblockN{Dewang Agarwal}
\IEEEauthorblockA{\textit{On Device AI} \\
\textit{Samsung R\&D Bangalore}\\
Bangalore, India \\
dewa.agarwal@samsung.com}
\and
\IEEEauthorblockN{Siva Prasad Thota}
\IEEEauthorblockA{\textit{On Device AI} \\
\textit{Samsung R\&D Bangalore}\\
Bangalore, India \\
siva.prasad@samsung.com}
}

\maketitle

\begin{abstract}
With the advent of internet, not safe for work (NSFW) content moderation is a major problem today. Since, smartphones are now part of daily life of billions of people, it becomes even more important to have a solution which could detect and suggest user about potential NSFW content present on their phone. In this paper we present a novel on-device solution for detecting NSFW images. In addition to conventional pornographic content moderation, we have also included semi-nude content moderation as it is still NSFW in a large demography. We have curated a dataset comprising of three major categories, namely nude, semi-nude and safe images. We have created an ensemble of object detector and classifier for filtering of nude and semi-nude contents. The solution provides unsafe body part annotations along with identification of semi-nude images. We extensively tested our proposed solution on several public dataset and also on our custom dataset. The model achieves F1 score of 0.91 with 95\% precision and 88\% recall on our custom NSFW\_16k dataset and 0.92 MAP on NPDI dataset. Moreover it achieves average 0.002 false positive rate on a collection of safe image open datasets.
\end{abstract}


\section{Introduction}
\begin{figure*}[htbp]
\centerline{\includegraphics[scale=0.75, trim={0 6cm 3cm 0},clip]{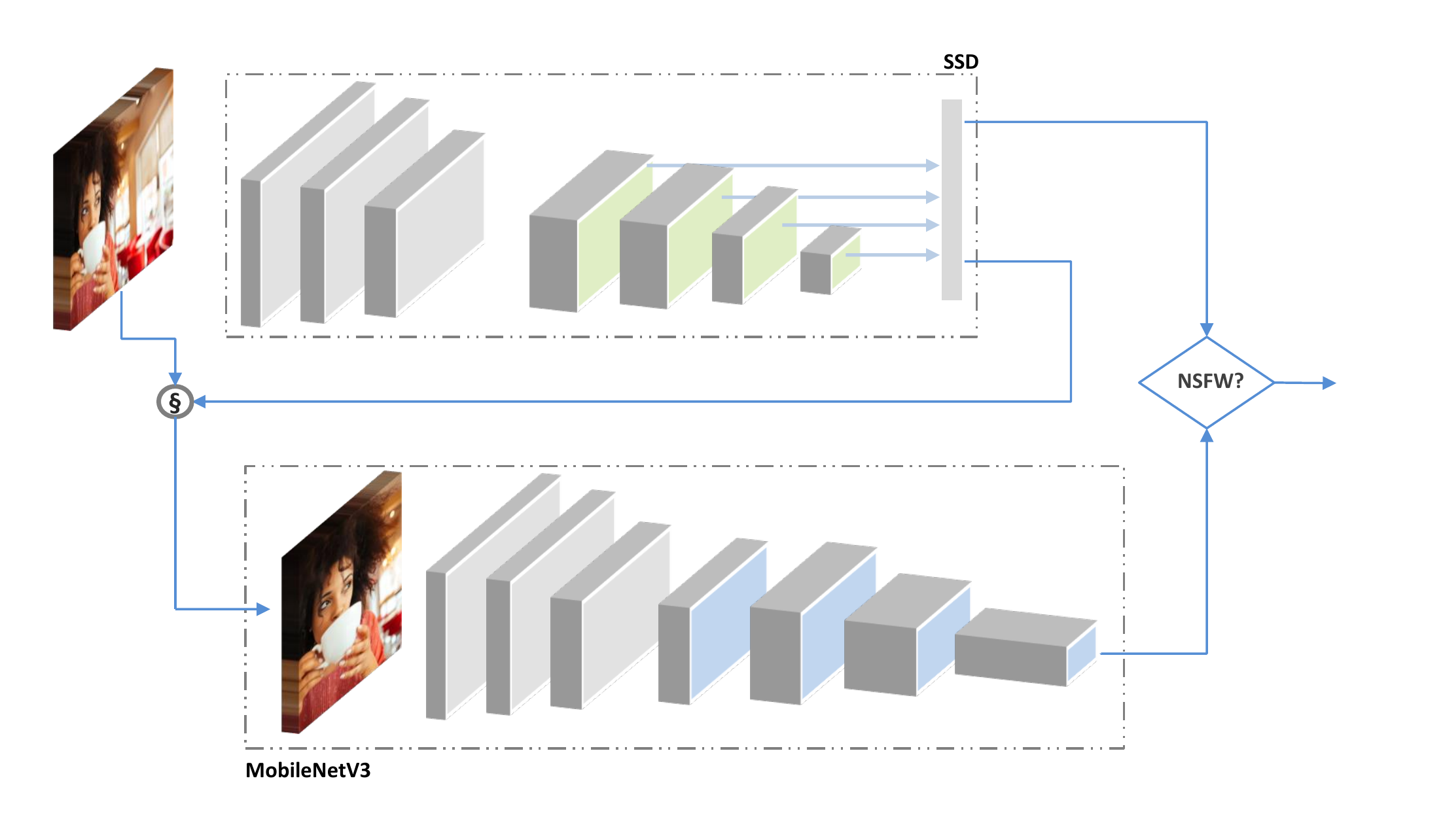}}
\caption{The diagram shows the proposed ensemble. Image is first resized and passed to unsafe body part detector which provides unsafe parts and human localized blocks. Then these localized blocks are resized and passed through the classifier. Here complete image is passed to the classifier if more than 2 distinct people are detected.}
\label{fig1}
\end{figure*}

In current digital age, mobile phones have become an integral part of young people and majority of children now use phones. With the widespread usage of social media, billions of images are exchanged everyday. Due to abuse of such platforms, teenagers are exposed to suggestive and explicit images. Pornographic contents have psychological as well as physical consequences across all ages. According to a study pornographic addiction leads to brain volume loss \cite{b1}. Depression and lack of interest is also common in case of prolonged exposure. Thus, making moderation of NSFW contents present on these devices much more important. 

An average smart phone user stores thousands of images and thus it becomes very cumbersome to manually filter out suggestive or explicit images. Considerable amount of research work has been done in creating systems to detect and filter out images with pornographic content but majority of these solutions are cloud based\cite{b2} and are not designed for mobile devices. This is still a very challenging problem \cite{b4} as in some cases only a portion of a private body part is visible while the complete image is safe for work (SFW) or the image is suggestive in nature or male breasts are SFW while female breasts are NSFW. Moreover, the usual vision problems still remains likes occlusion, blurring, small identifier, etc. 

Today as people are becoming more and more privacy oriented, they are reluctant in uploading their personal images along with few potential NSFW images on a server-side solutions to find NSFW images. Also there are certain level of regional biases seen in some of these solutions. In a study conducted by our team on around 210 people more than 190 people were interested in having an automated tagging feature on their phones which can detect private images and allow the user to remove or move them to a secure folder. Here private images included pornographic as well as semi-nude images e.g. bikini, lingerie, swimsuit, etc. This feature should work locally without uploading any information on server.

NSFW definition is subjective as it can change for an individual based on various factors like demographic, ethnicity, personal preferences, etc. For this study we consider nude, suggestive, semi-nude(includes lingerie, bikini, etc.) images as NSFW. This definition excludes animated and god images. Due to lack of data pertaining to our definition of NSFW, a new data set was created containing nude images with body part(female breasts, female genitalia, male genitalia, buttocks and person) annotations and semi-nude images. For safe dataset we collected some controversial categories like gods, kids, selfies while including images from some common reoccurring classes. This complete data set was used to train an ensemble of models for detection of such NSFW content on-device. We have experimented with different dataset and model variations. In subsequent sections we present in depth about literature review, methodologies used, and discuss about our experiments and their results.

\section{Literature Review}
In the past decade a lot of work has been done on identifying pornographic images from rest. Traditionally skin-based methods were popular followed by few studies based on hand crafted feature to mitigate some of the problems with  skin-based methods. Later on with the success of Convolutional Neural Network (CNN) in field of vision different neural feature based methods have been proposed.  
\subsection{Skin Detection Based Methods}Initially multiple skin detection based methods have been proposed \cite{b5}, \cite{b6}, \cite{b7}, \cite{b8}, \cite{b20}. In these studies, skin regions were extracted and then they were classified for adult content based on a trained classifier or by the amount of exposure.  In \cite{b8}, images are filtered by a skin color model to extract the skin regions and then they are fed into an support vector machine for binary classification. The main problem with these methods is that skin detection itself is a complex task. Also cases like that of a selfie in which face occupies considerable portion of the image or cases with skin colored regions(be it clothes or animals) creates further complications for such systems. 

In \cite{b24} authors provided improvement over \cite{b36} by extracting similar features of the same skin regions and using detected faces as features to mitigate false positives. Finally they trained different ML classifiers to distinguish between pornographic and bikini images and achieved best results with random forest approach.
 
\subsection{Hand Crafted Feature Based}
To overcome some of the skin detection based limitations, many researchers utilized the visual Bag-of-Visual-Words (BoVW) to extract the feature for recognizing pornographic images \cite{b9}, \cite{b10}, \cite{b12}, \cite{b17}. In \cite{b10} a BoVF approach based on the HueSIFT\cite{b11} is applied to still images classification between nude and non-nude images. Because these are hand crafted features they fail to generalise well on broader problem.

Authors in \cite{b13} designed an algorithm which uses a combination of Daubechies' wavelets, normalized central moments, and color histograms to provide semantically-meaningful feature vector matching to provide comparisons between a query image and images in a pre-marked training set efficiently and effectively. 

\subsection{Neural Feature Based}
There have been numerous region of interest (ROI) based methods \cite{b14}, \cite{b18}, \cite{b19}, \cite{b21}, \cite{b22}, \cite{b23} using neural features for training a classifier. Moving from traditional methods in \cite{b14} author experimented with AlexNet \cite{b15} based classifier, GoogLeNet \cite{b16} based classifier and an ensemble of the two. GoogLeNet based classifier showed significantly better results than AlexNet based classifier because of the deeper architecture while the ensemble showed slightly improved results over GooLeNet based classifier alone.

In \cite{b23} authors used weighted multi instance learning to address local region based detection of private body parts. For training the image is divided into a bag of images with overlapping portions with key pornographic portion. As per the study, this aided the model to learn key context present in positive images. For evaluation the image is divided into fixed 5 images parts which are resized and fed to model for evaluation. But multiple evaluation for a single image on-device adds up on run time, memory and power consumption. Moreover, key annotations can not be created for semi-nude images or images with suggestive nudity with no visible private parts.

The biggest drawbacks with most of these approaches is that they solely target pornographic content. Case of semi-nudity which is inappropriate in some demography still remains largely untouched and there is a lack of training as well as testing data for our problem statement.

\section{Methodology}

In order to provide an on-device solution for detecting NSFW images, our solution uses deep learning adhering to recent success of deep learning in learning useful representations of visual data. Given the need to find semi-nudity along with key pornographic content in image, we propose an ensemble which consists of Single Shot MultiBox Detector(SSD) \cite{b25} with MobileNetV3\cite{b26} as feature extractor and a MobileNetV3 classifier. In this pipeline, the image is first passed through SSD to detect any NSFW body part and then human localised portion from image is passed down to the classifier for detection of NSFW content. With SSD we get human localisation and unsafe body parts present in the image while the classifier is responsible for detecting partial nudity content(lingerie, swimwear, etc.) and it also detects nudity content. This approach allows to address some of the drawbacks of solutions in literature(far field objects, natural adversaries, etc.) while providing a small and efficient on-device solution for the problem. In next few subsections we discuss in detail about the different parts of ensemble.

\subsection{Bodypart Detector}
In order to capture the unsafe bodyparts present in an image we use object detection. The same object detector also provides human localization portion of the image which is fed to the classifier for further predictions if needed. Many object detectors have been proposed in the past like Faster-RCNN\cite{b32}, SSD, YOLO\cite{b34}, R-FCN\cite{b33}. In SSD the image is passed through a standard feature extractor once to compute feature map. Then object scores are calculated using a small convolution filter. In order to deal with objects of different scales multiple features maps are used in succession to initial feature map using convolutions. This way SSD achieves decent latency vs accuracy trade-off. As per \cite{b30}, SSD with MobileNet provides a faster model with relatively low execution time when comparing with some other methods. Thus, we use SSD with MobileNetV3 as feature extractor for a lightweight detector.

\subsection{MobileNet}
In recent past there have been a trend of using deeper and complicated network for increasing model accuracy. However deploying these solutions on mobile device is not always feasible because of memory and time constraints. MobileNet family models addresses this problem to an extent by providing decent accuracy-latency trade off on mobile devices. They provide a lightweight and efficient architectures for classification, detection and segmentation tasks. 

MobileNetV1\cite{b27} was based on a streamlined architecture that used depthwise separable convolutions replacing convolutional layers to build light weight deep neural networks. MobileNetV2\cite{b28} introduced the inverted residual layer with linear bottleneck in order to make even more efficient layer structures by leveraging the low rank nature of the problem. MobileNetV3\cite{b26} used network architecture search with NetAdapt algorithm along with some network modification including last layer redesign and soft h-swish non-linearity for more efficient trade-off between accuracy and latency. It reduced latency by 20\% for classification task as compared to MobileNetV2.

\begin{table*}[!htbp]
\caption{Bodypart Data distribution}
\renewcommand{\arraystretch}{1.5}
\begin{center}
\begin{tabular}{|c|c|c|c|c|c|}
\hline
\textbf{Annotation} & F\_BREAST & F\_GENITALIA & M\_GENITALIA & BUTTOCK & PERSON  \\
\hline
\textbf{Count} & 45989 & 30103 & 42668 & 19177 & 81371 \\
\hline
\end{tabular}
\label{tab1}
\end{center}
\end{table*}

\begin{table}[!htbp]
\caption{Classification top 5 classes data distribution}
\renewcommand{\arraystretch}{1.5}
\begin{center}
\begin{tabular}[c]{|c|c|c|c|c|c|}
\hline
\textbf{Class} & nsfw & person & chair & car & dining\_table \\
\hline
\textbf{Count} & 24915 & 24627 & 12671 & 12210 & 11784 \\
\hline
\end{tabular}
\label{tab2}
\end{center}
\end{table}

Initially we experimented with MobileNetV2 but found best performance with MobileNetV3. So, for the purpose of this study we have used MobileNetV3 pre-trained on ImageNet\cite{b29} for classification and MobileNetV3 pre-trained on COCO\cite{b31} dataset for bodypart detection. We apply transfer learning on this pre-trained model with our curated data to detect Nude as well as semi-nude content for classification task. For evaluation, localised image is resized and passed to the classifier for detecting the presence of NSFW content. Contrary to binary classifier used in most of the literature for NSFW detection, we created a multi-label classification using COCO train dataset for a diverse negative(safe) dataset. This helps the model in learning the required and beneficial representation in the image which reduces the false positive count to a very large extent. The impact of multi-label and multi-class classification is presented in the next section. We train the classifier with a modified weighted binary cross-entropy loss \eqref{eq} where we can adjust the precision and recall for required classes. 

\begin{equation}
L_{i}=-(\alpha_{i} y_{i} \log(p(y_{i})) + \beta_{i} (1-y_{i})\log(1-p(y_{i}))) \label{eq}
\end{equation}

where $\alpha_{i}$ and $\beta_{i}$ are the weight terms applied to change the precision and recall of the model, $y_{i}$ and $p(y_{i})$ are the true label and predicted label respectively.

\section{Experiments \& Results} 

In this section, we discuss about the dataset, training parameters and results of our proposed ensemble. At first we discuss about our dataset collection and sanitation process which includes nudity, semi-nudity and safe images. Next, we provide validation results of our proposed method on a separate testing dataset which we collected to make it closer to a regular phone user data along with a standard dataset.

\subsection{Dataset}

Due to lack of nudity and semi-nudity data and problem of copyrights with available data, we created a dataset containing safe as well as unsafe images containing nude and semi-nude images. We collected a total of 21,031 semi-nude images and 85,069 nude images. All the nude images were annotated for 5 annotations: female breasts, female genitalia, male genitalia, buttocks and person. Semi-nude images included images from categories like swimwear, bikini, back or side view nudity, etc. All these annotated images and the semi-nude images were reviewed iteratively to remove images with wrong annotations or false positives. This collected dataset maintains a wide variety of images from professional images of very high quality to non-professional images like selfies. The dataset also has decent distribution of different ethnicity to avoid any such bias.

For safe images, we used COCO train\_2017 dataset because of its diverse categories which helps the overall method to recognise the nude or semi-nude content better. It contains classes like bed, couch, cellphones, different animals and some other reoccurring categories. Moreover we aimed to avoid any false positive for controversial categories like gods and kids, so we collected images for the same. Next, we included a collection images for some categories like paintings, selfies, which could possibly give false positives. So we aggregated god, kids, paintings, selfies, and COCO dataset for our safe dataset collection. We cleaned the safe dataset for possible unsafe images according to our definitions in an iterative manner. This iterative cleaning was done with models trained on subsequent releases of annotated and reviewed dataset.

For training different components of our ensemble we distributed our data into two parts: \textit{bdprt\_data}, \textit{clf\_data}. The \textit{bdprt\_data} comprises of all the bodypart annotated dataset along with some portion of safe data.  Table.~\ref{tab1} shows the distribution of data in \textit{bdprt\_data}. This part of the dataset is used for training the bodypart detector. Here we found that by adding negative images(safe images) to the training dataset of SSD improves the precision of our resulting model as it reduces the false positive count. The \textit{clf\_data} contains full safe data and NSFW data which includes a subset of images from bodypart annotated data and semi-nude data. This resulting dataset is used for training the classifier which further provides confidence for NSFW content. 

We have experimented with two variations of \textit{clf\_data}: \textit{clf\_data\_full} and \textit{clf\_data\_two}. In \textit{clf\_data\_full}, we include all the classes from COCO train\_2017 dataset along with a NSFW class. Thus the total class count is 81(80 from COCO and 1 NSFW class). Here the person class contains safe person images while the added NSFW class contains above mentions NSFW images. Also the NSFW class images are added to original COCO classes with multiple re-iterative training. We created \textit{clf\_data\_two} dataset for binary classification task containing NSFW and SFW classes only. The NSFW class is same as that for \textit{clf\_data\_full} while for SFW class we take images from each of the 80 classes to get SFW class. The final distribution of top 5 classes in \textit{clf\_data\_full} is shown in Table.~\ref{tab2}. 

\begin{figure}[ht!]
   \centering
  \subfloat[]{%
       \includegraphics[width=0.3\linewidth, height=0.3\linewidth]{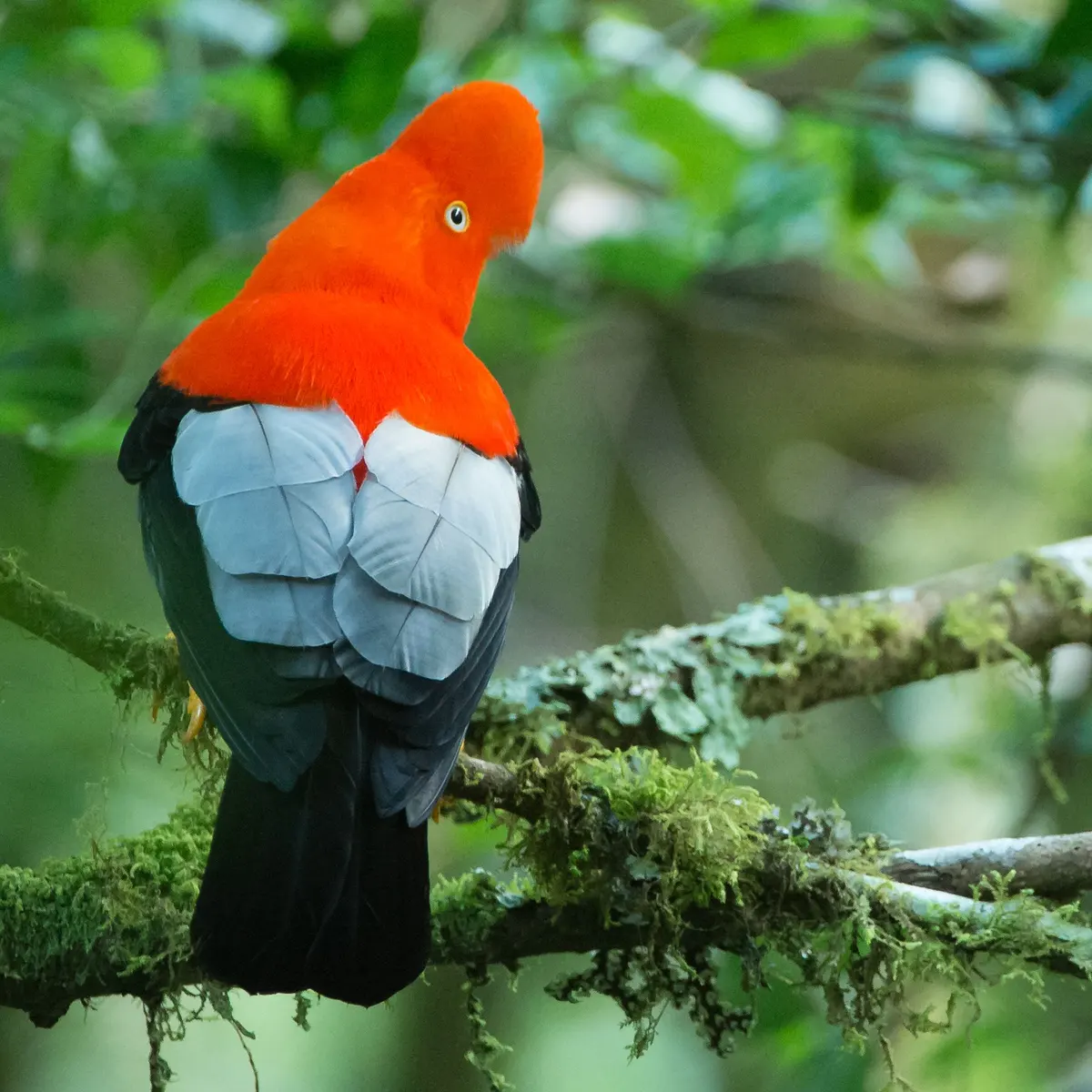}}
    \hfill
  \subfloat[]{%
        \includegraphics[width=0.3\linewidth, height=0.3\linewidth]{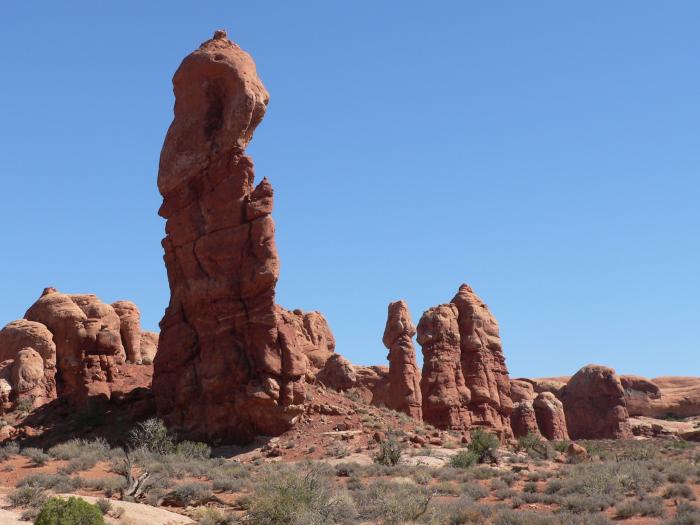}} \hfill
  \subfloat[]{%
        \includegraphics[width=0.3\linewidth, height=0.3\linewidth]{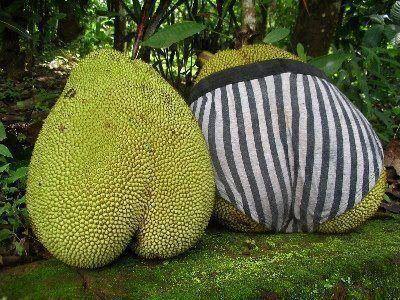}}
\\
  \subfloat[]{%
        \includegraphics[width=0.3\linewidth, height=0.3\linewidth]{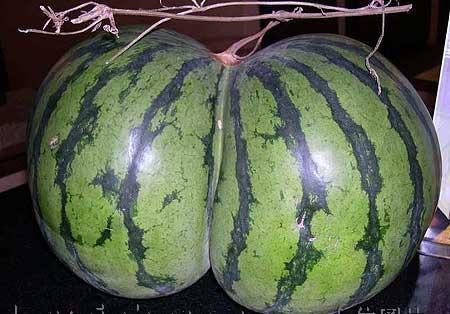}} \hfill
\subfloat[]{%
        \includegraphics[width=0.3\linewidth, height=0.3\linewidth]{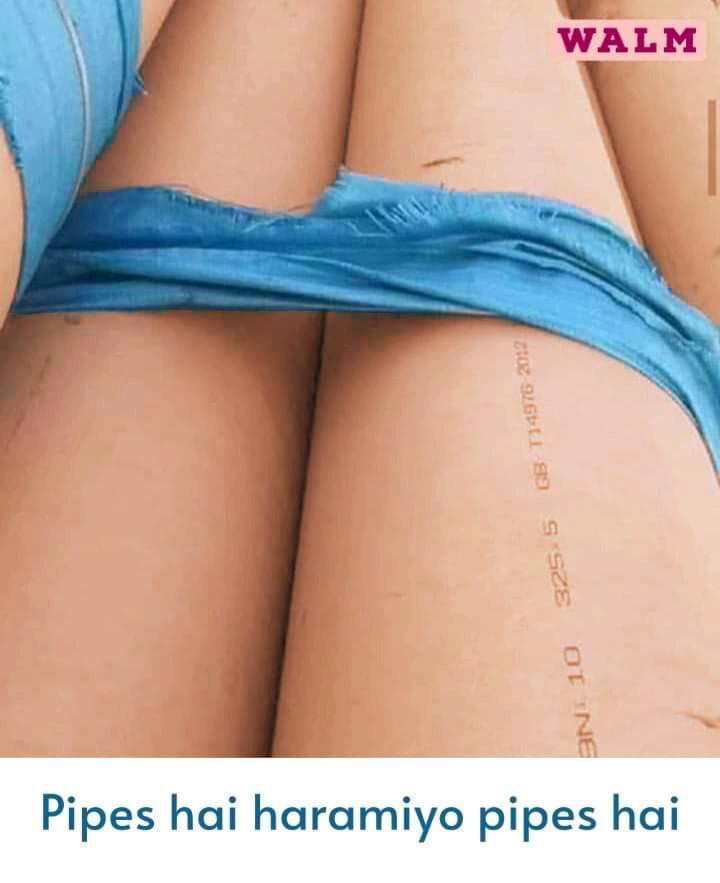}} \hfill
\subfloat[]{%
        \includegraphics[width=0.3\linewidth, height=0.3\linewidth]{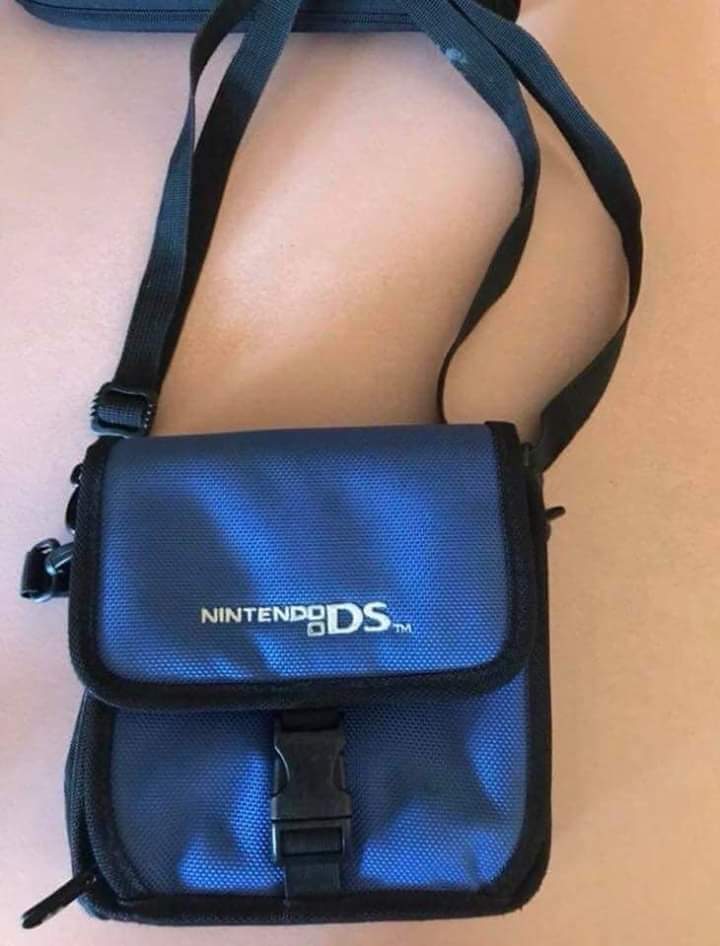}} 
  \caption{(a) .. (f) Some natural adversarial examples from created testing data. All these images have visual similarities to unsafe bodyparts and are the limitation of traditional approaches.}
  \label{fig2} 
\end{figure}

\begin{table*}[!ht]
\caption{Performance on NSFW\_16K}
\renewcommand{\arraystretch}{1.5}
\begin{center}
\begin{tabular}[c]{lccccccc}
\hline
Models & TP Count & TN Count & FP Count & FN Count & Precision & Recall & F1 \\
\hline
\textbf{Ensemble1} [\textit{clf\_data\_full}] (\textbf{Proposed}) & 4586 & 10714 & 239 & 623 & 0.95 & \textbf{0.88} & \textbf{0.914} \\
Ensemble2 [\textit{clf\_data\_two}] & 4588 & 10568 & 385 & 621 & 0.92 & 0.88 & 0.90 \\
Ensemble1 w/o Localization & 4127 & 10775 & 178 & 1082 & 0.96 & 0.79 & 0.87\\
Ensemble1 w/o Aug & 4344 & 10769 & 184 & 865 & 0.96 & 0.83 & 0.89 \\
Ensemble1 [$\alpha_{i}=1$, $\beta_{i}=1$] & 4600 & 10540 & 413 & 609 & 0.92 & 0.88 & 0.90 \\
Ensemble1 [$\beta_{nsfw}=2$] & 4162 & 10876 & 77 & 1047 & \textbf{0.98} & 0.80 & 0.88 \\
OpenYahoo & 1241 & 10854 & 99 & 3968 & 0.93 & 0.24 & 0.38 \\
\hline
\end{tabular}
\label{tab3}
\end{center}
\end{table*}

\begin{table*}[!ht]
\caption{Performance on different public dataset}
\renewcommand{\arraystretch}{1.5}
\begin{center}
\begin{tabular}[c]{lccccccc}
\hline
\multirow{2}{*}{\text{Metrics }} & \multicolumn{4}{c}{\text{Dataset}} \\
 & Caltech-256 Test\cite{b38} & Coil-100\cite{b39} & Places-365 Test\cite{b40} & Cumulative \\
\hline
Total Image Count & 30607 & 7200 & 327774 & 365581\\
False Positive Count & 67 & 0 & 726 & 793 \\
False Positive Rate & \textbf{0.0021} & \textbf{0} & \textbf{0.0022} & \textbf{0.0022} \\
\hline
\end{tabular}
\label{tab4}
\end{center}
\end{table*}

\textbf{NPDI dataset\cite{b35}}: The dataset consists around 80 hours of videos of 40 pornographic and 40 non-pornographic videos. All these videos are segmented into a total of 16,727 images out of which images are pornographic, 333 are hard-non pornographic images and remaining 333 images are easy non-pornographic images. The hard non pornographic images are collected via searching images of categories like swimming, wrestling, beach which coincides with our NSFW definition so because of that we remove the hard non-porn images. Further we loosely reviewed the porn images and removed false positives as the key segments chosen for creating this dataset is not optimal. In addition, we also removed all the animated images from dataset as it was out of scope for our solution. Finally we prepared a reviewed NPDI dataset which consists of 6785 safe and 3803 unsafe images. One major drawback which we notice about this dataset is that the image resolution is very poor with majority of the images have dimensions below 300.

\textbf{NSFW\_16k}: We have curated a separate test dataset by collecting several user data. The main aim of this dataset(NSFW\_16k) is to make it closer to a smartphone user data while making it diverse. Majority of the user data which we collected was safe so we had to add NSFW data in it. We added separate NSFW content  from the training set while making it as varied as our training set. We also added natural adversarial examples in this dataset which have certain visual similarities to nude or semi-nude content but were otherwise safe natural images. Some such examples are shown in Fig.~\ref{fig2}. In total the testing dataset comprises of in total 16142 images out of which 10933 images are safe images while 5209 images are NSFW. All these unsafe images contains both nude and semi-nude images.

\subsection{Training and Network Parameters}

We trained our model in a greedy manner by training the bodypart detector and classifier separately. Here we used separate training and validation dataset for training both the components and train them with full image. We use mix of rotation, random crop and random Gaussian blur for training time data augmentation.

For training the SSD we use pre-trained weights from COCO dataset and train them on \textit{bdprt\_data}.  For training and evaluation, images are resized to 300x300 while preserving aspect ratio. We used RMSProp optimizer, batch size of 32, and starting learning rate of 0.004. Moreover we do training time quantization which allows compression of final resulting model which is to be deployed on-device. 

For training the classifier we use MobileNetV3 pre-trained on ImageNet and do transfer learning by changing the final layer from 1000 classes to 81 classes. The input image is resized to 224x224 with anti-aliasing and we use SGD with Nestrov momentum with initial learning rate of 0.04, momentum of 0.9 with batch size of 32. Here also we do training time quantization allowing us to create quantized model with integer operation support instead of float. We used weighted cross entropy loss here with Sigmoid activation in last layer when doing multi-label training with clf\_data\_full and used Softmax for clf\_data\_two. When using other adaptive optimization methods we found that they failed to converge. This can be because of poor generalization capability of adaptive optimization methods \cite{b37}.

\subsection{Results}

Table.~\ref{tab3} shows the performance of our proposed solution evaluated on NSFW\_16k dataset. Here Ensemble 1 is trained on \textit{clf\_data\_full} with $\alpha_{nsfw}=2$
, $\beta_{nsfw}=2$ with localization and augmentation. The proposed ensemble trained with \textit{clf\_data\_full} performs better when compared with ensemble trained with \textit{clf\_data\_two}. Also we see performance drop without human localization. This is in line with expected result that far-field objects are better captured if a localised crop of the object is evaluated. Moreover doing human localization along with augmentation allows the model to avoid significant amount of natural adversaries Fig.~\ref{fig2} while allowing for better overall results. Moreover, we can note the better results when compared with ensemble trained without augmentation or without weights or without localization. In Table.~\ref{tab3} we can note our proposed solution has the best trade-off between precision and recall while the drop in precision is far less when compared with best precision model Ensemble1 [$\beta_{nsfw}=2$]. Also here we can see that OpenYahoo perform poorly and this is because of the limitation of semi-content, the solution couldn't differentiate well between semi-nude and other images. Moreover the performance of OpenYahoo is sub-par even for nudity content.

In a consumer aware market we need to have a high precision solution. This is in conscious that an average smartphone user stores thousands of images majority of which are safe. Here a 5\% decrease in precision from 95 to 90 can lead to hundreds of extra images for user to review before moving or deleting the suggestions. Thus to find out the feasibility of our solution we evaluated the proposed pipeline on some public dataset containing safe images mainly and the results are shown in Table.~\ref{tab4}. Here we can note the robustness of our proposed ensemble provided with safe images. The average false positive rate is around 0.002 which is almost insignificant. While reviewing the few false positives we note that almost all such images had a person present them. Some cases in these false positives can be argued if they are NSFW or not. The proposed solution mitigates false positives for wide categories of objects while showing false positives rarely in case of person class(images with person present in them) only.

\begin{table}[!t]
\caption{Comparison of MAP on NPDI dataset}
\renewcommand{\arraystretch}{1.5}
\begin{center}
\begin{tabular}[c]{lc}
\hline
Methods & MAP \\
\hline
Deep Region-based CNN\cite{b23} & 0.87  \\
Deep Part Detector\cite{b23} & 0.87  \\
Deep MIL\cite{b23} & 0.86  \\
Weighted MIL\cite{b23} & \textbf{0.97}  \\
\hline
OpenYahoo\cite{b3} & 0.79 \\
\hline
Ours [\textit{clf\_data\_two}] & 0.93 \\
Ours [\textit{clf\_data\_full}] & 0.92 \\
\hline
\end{tabular}
\label{tab5}
\end{center}
\end{table}

We evaluated the proposed solution on reviewed NPDI dataset and Table.~\ref{tab5} shows the performance of the solution. Here we have taken top 4 rows from \cite{b23} as we were not able to reproduce their results due to dataset constraints. When comparing with other results on this dataset, our method performs fairly and achieves MAP of 0.93. This is when our method is using smaller models with real time inference on mobile devices. Upon further inspection of results we found that some performance drop is seen because the dataset contains a large amount of images with resolution less than 300x300 and when these images are up-scaled for evaluation, the model results degrade. This is possible as our dataset contains mostly higher resolution images and not many up-scaled images are seen in training and thus our proposed solution has some performance drop when evaluating such images.

We also carried out user trials with our solutions on smartphones. We received positive feed backs with less that 5\% false detection while enriching overall user experience. The average inference time of our proposed solution on Samsung S20 device comes out to be around 85 milliseconds for one image. Individually, bodypart detector takes around 60 milliseconds while classifier takes 25 milliseconds. The inference is done using device's CPU.

\section{Conclusion}
In this paper, we have proposed a deep learning solution for detecting nudity and semi-nudity contents. We have curated a dataset containing diverse set of safe and unsafe images including semi-nudity images for training and evaluation of our solution. For this we proposed a deep learning ensemble containing MobileNetV3 classifier and SSD with MobileNetV3 feature extractor. Here SSD detects unsafe bodyparts while also providing human localized portion of image which is then fed to the classifier for further classification.  We have shown that our proposed method on-device performs on par with other pornographic detection solutions while detecting semi-nude content as well. It provides the best trade-off between accuracy and recall while maintaining a higher precision. Our proposed method gives F1 of 0.91 on our custom testing dataset with precision of 0.95. We extensively tested our solution on safe public dataset and showed that it achieved less than 0.2 false positives percentage. This provides a lightweight solution which can be deployed on-device with inference time of 85 milliseconds on-device.

\end{document}